\documentclass[12pt]{article}

\usepackage{graphicx}
\usepackage{amssymb}
\usepackage{amsthm}
\usepackage{booktabs}
\usepackage{rotating}
\usepackage{multirow}
\usepackage{eurosym}
\usepackage{amsfonts}
\usepackage{amsmath}
\newtheorem{definition}{Definition}
\newtheorem{assumption}{Assumption}
\usepackage{mathtools} 
\usepackage{tabularx}
\usepackage{umoline}
\usepackage{caption}
\usepackage{subcaption}
\usepackage{color}
\usepackage{url}
\usepackage{authblk}
\usepackage[english]{babel}

\usepackage{pbox}
\usepackage{footnote}
\usepackage{multirow}
\usepackage{changepage}
\usepackage{tablefootnote}
\usepackage{xcolor}

\usepackage[margin = 2cm]{geometry}
\usepackage{booktabs}
\usepackage{adjustbox} 
\usepackage{multirow} 
\usepackage{subcaption, epsfig} 
\usepackage{graphicx}
\usepackage{subcaption}
\usepackage{hyperref}

\usepackage{breakurl}
\usepackage{hyperref}
\hypersetup{breaklinks=true}

\usepackage[natbibapa]{apacite}

\usepackage{rotating}
\usepackage{lscape}
\usepackage{float}
\floatstyle{plaintop}
\restylefloat{table} 
\usepackage{booktabs}
\usepackage{textcomp}
\usepackage{siunitx}
\usepackage{multirow}
\usepackage{makecell}
\usepackage{longtable}
\usepackage{setspace}
\usepackage{amssymb}
\usepackage{amsmath}
\usepackage{booktabs}
\usepackage{adjustbox} 
\usepackage{multirow} 
\usepackage{subcaption, epsfig} 
\usepackage{graphicx}
\usepackage{subcaption}
\usepackage{hyperref}
\usepackage{utfsym}
\usepackage{multicol}
\usepackage{rotating}

\usepackage{array}
\usepackage{natbib, ragged2e}
\usepackage{appendix}
\usepackage{pdflscape}
\usepackage{adjustbox}

\begin{document}
\newcolumntype{B}[1]{>{\small\hspace{0pt}\RaggedRight\bfseries}p{#1cm}}
\newcolumntype{C}[1]{>{\RaggedRight}p{#1cm}}

\title{Are causal effect estimations enough for optimal recommendations under multitreatment scenarios?\footnote{\scriptsize NOTICE: This is the author´s version of a submitted work for publication. Changes resulting from the publishing process, such as editing, corrections, structural formatting, and other quality control mechanisms may not be reflected in this document. Changes may have been made to this work since it was submitted for publication. This work is made available
under a Creative Commons BY license. \textcopyright CC-BY-NC-ND}}

\author[1, 2]{Sherly Alfonso-Sánchez}
\author[1]{Kristina P. Sendova}
\author[1]{Cristi\'{a}n Bravo}

\affil[1]{Department of Statistical and Actuarial Sciences, Western University, 1151 Richmond Street, London, Ontario, N6A 5B7, Canada.}
\affil[2]{Departamento de Matemáticas, Universidad Nacional de Colombia. Ave Cra 30 \#45-03, Edificio 404. Bogotá, Colombia.}

\date{}
\maketitle
\begin{abstract}
When making treatment selection decisions, it is essential to include a causal effect estimation analysis to compare potential outcomes under different treatments or controls, assisting in optimal selection. However, merely estimating individual treatment effects may not suffice for truly optimal decisions. Our study addressed this issue by incorporating additional criteria, such as the estimations' uncertainty, measured by the conditional value-at-risk, commonly used in portfolio and insurance management. For continuous outcomes observable before and after treatment, we incorporated a specific prediction condition. We prioritized treatments that could yield optimal treatment effect results and lead to post-treatment outcomes more desirable than pretreatment levels, with the latter condition being called the prediction criterion. With these considerations, we propose a comprehensive methodology for multitreatment selection.
Our approach ensures satisfaction of the overlap assumption, crucial for comparing outcomes for treated and control groups, by training propensity score models as a preliminary step before employing traditional causal models. To illustrate a practical application of our methodology, we applied it to the credit card limit adjustment problem. Analyzing a fintech company's historical data, we found that relying solely on counterfactual predictions was inadequate for appropriate credit line modifications. Incorporating our proposed additional criteria significantly enhanced policy performance.
\end{abstract}

\begin{keywords}
Causal Learning, Conditional Value-at-Risk, Credit-limit adjustment, Counterfactual Prediction.
\end{keywords}

\section{Introduction}
\label{sec:intro}
With advancements in data science, numerous new applications have emerged in various fields, including e-commerce, finance and banking, healthcare, marketing and advertising, and insurance. For example, \citet{knuth2022consumer} demonstrated that the use of a decision tree as a data mining technique adequately identified several patterns of fraud that could improve the identification of fraudulent online shopping transactions, and \citet{korangi2023transformer} showed that a transformer-based model with a loss function for multi-label classification and multichannel architecture outperformed traditional models in the task of predicting default term structure over the short-to-medium term for mid-cap companies (i.e., publicly listed firms with a market capitalization below 10 billion USD). \citet{singh2019automating} have proposed an end-to-end system to automate the process of registration, processing and deciding regarding the final claim in a car accident, improving the waiting times for insurance companies and policyholders; indeed, this system uses deep learning techniques using images of the vehicle after the collision as inputs. All of these developments are predictive rather than causal, as they do not state what the outcome would be under different circumstances. 

Unfortunately, several predictive models have been mistakenly adopted to generate causal effects, as stated in \citep{ramspek2021prediction}. However, they may not offer causal interpretations but rather the understanding of correlations or associations between the inputs and outputs. That is the reason why different researchers have advocated for causal analysis to enhance decision-making. For example, \cite{cankaya2023evidence} utilized machine learning and Bayesian Belief Networks to develop a decision support system that helps decision-makers identify causal relationships and mitigate risks in aviation operation. Similarly, \cite{liu2022optimization} employed a novel causal approach, integrating Causal Bayesian Networks (CBNs) with mathematical programming, to optimize interventions and improve supply chain resilience during disruptive events like the COVID-19 pandemic. In the same vein, \cite{liu2023robust} used CBNs and robust optimization to model and minimize disruption risks in the worst case scenarios, aiming to support decision-makers in selecting the best intervention strategies to maintain supply chain stable during uncertain events. Moreover, according to \citet{prosperi2020causal}, the premise that predictive algorithms lead to better decision-making processes has been questioned  mainly for scenarios in which interventions should be determined. For instance, a predictive model that identifies customers with a high probability of default cannot answer the question of what is the best loan-restructuring option. 

In many fields, determining how an outcome is perturbed by an action, that is, the treatment effect of an action, is an essential question. Moreover, with rapidly emerging data sources and, more recently, new data-science and statistical techniques, researchers and policymakers have exhibited increasing interest and need regarding the estimation of heterogeneous treatment effects concerning individual features,  rather than focusing solely on average causal effects. For example, a cancer patient would be more interested in knowing the individual causal effect of a treatment given that they have a more-aggressive tumor or a particular gene mutation, in place of knowing the average treatment effect in the population to which they belong \citep{bellon2015personalized}. In addition, in marketing campaigns, it is desirable to target persuadable customers to reduce costs and avoid irritating customers who do not wish to receive any unsolicited advertisement, also known as \textit{sleeping-dog clients} \citep{rzepakowski2012uplift}.

Indeed, the estimation of heterogeneous treatment effects is of concern in several areas that aim to create of adequate personalized recommendations, such as the case of education \citep{lehrer2022multiple}, healthcare \citep[][]{fan2022estimation,goradia2024made}, 
marketing \citep[][]{ba2022mobile,lei2022swayed,li2016matching}, economics \citep{baumann2021estimating}, and corporate governance \citep{lu2015covariate}. Banking is not outside this umbrella, as counterfactual thinking should be present for establishing a decision-making process that impacts the future behavior of customers. This is particularly the case for credit card portfolios, given that credit card payments have replaced cash transactions due to the impact of digital commerce and the COVID-19 pandemic \citep{cherif2022credit}. Consequently, studying how new credit policies can impact business profits and losses is of interest to both practitioners and academic investigators.

One of these possible policy changes regarding credit card portfolios is the determination of how credit limits should be adjusted over time, updating them according to the past behavior of a customer. Certainly, credit lines affect the manner in which customers spend. In this matter, \cite{aydin2022consumption} found that credit limit increases had a meaningful repercussion in the use of credit. Although the problem of credit limit adjustment is essential in any company that offers credit card products, it has not been widely studied in contrast to the determination of the default probability (e.g., \cite{djeundje2019dynamic}, \cite{sariannidis2020default}, and \cite{ALONSOROBISCO2022102372}). Furthermore, the counterfactual prediction framework for this specific problem has not been explored beyond theoretical developments. 

In this report, we present a comprehensive methodology for selecting treatments in scenarios in which treatment options vary continuously (i.e., different treatment dosages) and the outcome is also continuous, with observability both before and after treatment (Our method is generic and can be adapted to situations in which the desirable outcome either increases or decreases. For simplicity, in this work we have assumed that a higher outcome is related to a more desirable outcome, without any loss of generality). Specifically, we transformed the problem into one with discrete treatment options, focusing on satisfying the crucial overlap assumption necessary for accurate outcome comparison between treatment and control groups. Additionally, we investigated the impact of policy performance by incorporating an adaptation of the conditional value-at-risk (CVaR) concept, commonly used in portfolio management, to measure the uncertainty of impact predictions in a causal framework and reduce the overall risk of an incorrect decision. Furthermore, we evaluated the inclusion of a predictive condition in these criteria, namely prioritization of treatments that not only could achieve optimal results in terms of treatment efficacy but also could lead to post-treatment outcomes surpassing those observed before treatment initiation. To showcase the usefulness of our methodology, we have addressed the credit card limit-setting problem, treating it as a discrete treatment-dosage problem, and have contributed to the limited existing literature regarding the development of methods for credit line modification that incorporate counterfactual learning. We aimed to address whether this unique criterion alone was sufficient to produce an optimal solution. To answer this question, we utilized financial data from a real-world credit card portfolio. The following research questions guided our investigation in the case of a multitreatment scenario:  
\begin{enumerate}
    \item[\textbf{RQ1:}] \textbf{Can well-known methods for measurement of estimation uncertainty in portfolio management be used in addition to the recommendation generated by causal effect estimations? }
    \item[\textbf{RQ2:}] \textbf{How can an appropriate prediction condition be incorporated into the recommendation of treatment using causal inference predictions and uncertainty measurement, considering the existence of observable pre- and post-treatment outcomes?}
    \item[\textbf{RQ3:}] \textbf{Is counterfactual prediction sufficient to generate optimal suggestions for continuous treatments, specifically in financial scenarios such as credit card limit adjustments? Does the incorporation of uncertainty measures and/or the inclusion of an adequate prediction condition improve the final decision?}
\end{enumerate}
The remainder of this article is organized as follows: In Section 2, we present a literature review of related work. In Section 3, we introduce background concepts for the development of our research, such as the definition of individual treatment effects (ITEs), some traditional causal learning models  relevant to our research, and the procedure we used for causal model selection. The methodology for generating the recommendation of treatment dosage, together with the inclusion of the uncertainty assessment and prediction condition, is presented in Section 4. Finally, we outline the implementation of the proposed methodology for the particular objective of credit limit adjustment. The results of the applied strategies, as well as a discussion of our work and the conclusions derived from it, are presented in Sections 5, 6, and 7, respectively.

\section{Related Work}
To estimate the change in an outcome resulting from the change of an intervention (\textit{treatment}) given certain subject characteristics, two major approaches have been developed: \textit{uplift modeling} and \textit{treatment effect heterogeneity modeling}. The difference between these two research perspectives is that the former implicitly assumes that the data comes from a randomized experiment, whereas the methods developed for the latter explicitly expose the assumptions accepted for the use of experimental or observational data. Although investigations by both communities have appeared disjointed, \citet{zhang2021unified} have presented fundamental assumptions necessary for unifying both methods under the potential-outcome framework \citep{rubin1974estimating}. 

Uplift modeling has been explored to improve the success of policies in marketing campaigns, education, and business analytics, among others. For example, \cite{gubela2020response} introduced uplift models for maximizing incremental revenues rather than a traditional incremental sales target in marketing campaigns. They showed that revenue uplift modeling along with causal models can improve the profit of a campaign. \citet{olaya2020uplift} implemented uplift modeling to maximize the efficacy of retention efforts (e.g., the offering of tutorials) in higher education. They demonstrated the advantages of this strategy in comparison to the use of predictive models. In the financial industry, \citet{devriendt2021you} have introduced a novel evaluation metric, maximum profit uplift (MPU), capable of evaluating uplift models and prediction models of customer churn. With this measure, they established that uplift models outperformed predictive ones and therefore that uplift models could increase the profitability of retention campaigns. 

Research regarding treatment effect heterogeneity modeling has been focused on clinical data, and efforts have been concentrated on the use of real-world observational data, such as electronic health records, to gain information about heterogeneous patients and their responses to treatments, in contrast with the use of randomized control trial (RCT) data, which is the gold standard to introduce new drugs. However, RCT fails to generalize beyond the population of the study. Learning from observational data establishes challenges (e.g., the treatment selection bias because doctors do not select treatments randomly) and requires the acceptance of strong assumptions for the use of causal inference \citep{bica2021real}. Despite this condition, several models for the estimation of individualized treatment effects have arisen not only in the static scenario but also in the longitudinal setting, in which the treatment is effectuated several times; some examples of the first avenue are causal forest \citep{wager2018estimation}, two-model approach learners, X-learners \citep{kunzel2019metalearners}, R-learners \citep{nie2021quasi} and models based on deep learning (e.g., deep counterfactual networks with propensity dropout \citep{alaa2017deep}, generative adversarial networks for inference of individualized treatment effects [GANITE] \citep{yoon2018ganite}, Dragonnet \citep{shi2019adapting} and Hydranet  \citep{velasco2022hydranet}).  Regarding the longitudinal scenario, some relevant advances have been made by \citet{bica2020time} creating the Time Series Deconfounder method, which uses the assignment of multitreatments over time to estimate treatment effects in the presence of hidden confounders, and by \citet{melnychuk2022causal}, who have constructed the Causal Transformer, a transformer-based architecture for the estimation of counterfactual outcomes. All of these methodologies have been supported by relying on assumptions about the data and have been evaluated using synthetic or semisynthetic data, making their development difficult in practice.

Regarding causal inference in banking, \citet{kumar2022assessing} identified determinants to access the Kisan Credit Card program in rural India, and they also evaluated the impact of this strategy on the farm household incomes; specifically, they concluded that entry to the program had reduced farmers' dependency on moneylenders for borrowing credits by 25\%. \citet{kasaian2022effects} examined the effect of inclusion of a credit card ``teaser rate" (i.e.,  the initial interest rate offered to borrowers as a promotional rate at the beginning of a loan term), using information from a national bank in the United States. They deduced that these rates improve the generation of customer revenue by encouraging increased indebtedness. Those offers increase the customer's desire to borrow at regular interest rates that are higher than the teaser rate, a condition known as \textit{spillover-effect}. However, the use of causal inference for studying effects of credit card limits adjustment has been limited. In this avenue, \citet{miao2020intelligent} have presented a data-driven approach in which a balance response model was built to measure the heterogeneous treatment effect of different amounts of credit limit increase for different customers that had been divided into four principal groups depending on their credit rating, and a report by \citet{fahner2012estimating} proposes propensity score matching models to estimate counterfactual outcomes (balance 12 months after the decision was made to raise or maintain the credit limit), carefully examining the satisfaction of the overlap assumption to be able to generate reliable comparisons between treatment and control. In our research, we focused instead on constructing an expected profit response model because, with regard to the modification of current credit limits, not only the change in the balance but also the change in provisions, which are the expected losses due to default, should be considered. Additionally, we generated personalized recommendations rather than making decisions by groups of credit card holders and employed validation techniques for casual model selection that differed from supervised learning (in contrast with  \cite{miao2020intelligent} and \cite{fahner2012estimating}, respectively).

Although the generation of credit line modifications has not been extensively examined through the lens of causal learning, other researchers have advocated solutions to this problem using different techniques, although these decisions still rely on expert interventions and rule-based paths in practice.  For instance, \citet{dey2010credit} has presented the concept of using simulation with action-effect models to set the optimum credit limit per each account; however, this methodology was not validated with empirical results.  \citet{sohn2014optimization} used predictive models to determine the probability of default and the future balance, generating different groups of customers to be able to maximize net profit and selecting the ideal credit limit per group; nevertheless, the entire process requires manual interventions, and the proposed cost estimation ignores the total possible expected loss arising from the revolving nature of the credit card product.  \citet{so2011modelling} have suggested a dynamic credit-limit-setting policy generated from the use of behavioral scores and dynamic programming, but the cost of provisions is not explicitly considered in this case. \citet{charlier2019mqlv} have introduced a reinforcement learning (RL) algorithm called Modified Q-Learning for Vasicek (MQLV) to address optimal money-management policy for retail products based on clients’ aggregate financial transactions; however, they assume that the transactional data-sets satisfy the assumption of the Vasicek model, a condition that may be difficult to achieve in real-life situations. 
In addition, employing RL, \citet{alfonso2024optimizing} generated recommendations for increasing credit limits or keeping them constant using an off-line training strategy and a simulator to assess the impact of the actions (increase or maintain) according to historical information from a super-app. They aimed to optimize the expected profit of a company by considering the revenue and provision change, although they only considered binary treatment because of the limited amount of data. In this research, we focused on a more general treatment scenario in which the decision is suggested as a determined percentage of increase according to historical information. 

\section{Background}
In this section, we introduce some definitions and assumptions under the potential outcome framework for recommendations under multitreatment scenarios. In addition, some traditional causal models and a state-of-the-art methodology for selecting causal models are described.

\subsection{Definitions and Assumptions}
\label{subsec: Definitions and Assumptions}
\begin{definition}
We denote by $T$ a binary treatment variable, where $T=0$ represents having no treatment, that is, belonging to the control population, and $T=1$ for belonging to the treatment population. Under the potential outcome approach, there are two \textit{potential outcomes} for each individual $i$: $Y_i(0)$ is the potential outcome if the subject did not receive the treatment, and $Y_i(1)$ is the potential outcome if the subject received the treatment. For each individual $i$, the \textbf{individual treatment effect} (ITE) is defined as follows:
\begin{equation}
    ITE_i:=Y_i(1)-Y_i(0). \label{eqn:ITE_def}
\end{equation}
\end{definition}
Note that only one of the two potential outcomes is observable for a given participant; the unobserved potential outcome is called the \textit{counterfactual} of the observed outcome. In the following, for clarity, we have omitted the subscript $i$ on the $ITE$ variable.

It is possible to find the best estimator for the $ITE$, from observational data, provided that the following assumptions hold:

\begin{assumption}[Overlap]\label{assump:Overlap}
If $X$ denotes the vector of pre-treatment predictors for any individual with $X=x$, the probability of receiving the treatment and the probability of not receiving the treatment are each non-zero, expressed as follows:
\begin{equation}
    0< P(T=1 \vert X=x)< 1. \label{eqn:Propensity}
\end{equation}
\end{assumption}
The overlap assumption is required to guarantee that all types of subjects have been observed in the treatment and control groups, and based on this assumption be able to estimate the $ITE$. Many causal learning investigations have been supported by this assumption, which can be assured in the case of processing synthetic or semisynthetic data; however, with real-world data, it is imperative to validate this assumption through testing or identifying a subset of the predictor variables' domains for which this assumption holds true, ensuring the deduction of accurate conclusions.

\begin{assumption}[Stable unit treatment value (SUTV)] 
Every subject is independent of the others; that is, a treatment applied to an individual does not influence the outcomes for other individuals.
\end{assumption}
There are several scenarios in which the stable unit treatment value assumption (SUTVA), holds true. For example, when evaluating the efficacy of a new medication across different patients, we assume that each patient's response, such as survival probability, is independent of the others. Similarly, when analyzing changes in credit lines for various credit card holders, we assume that the expected profit is unaffected by the credit lines of other cardholders. However, in cases such as investigating churn probability for a car insurance policy, the introduction of a new customer service program aimed at expediting claim processing, which includes personalized assistance, can complicate matters. Positive experiences shared by some policyholders may influence the perceptions of others through word-of-mouth, potentially leading to increased satisfaction and faster times for claim processing.

\begin{assumption}[Unconfoundedness]\label{assump:Unconf}
The distribution of the treatment is independent of the distribution of potential outcomes, given the observed variables, expressed as follows:
\begin{equation}
    \left(Y(0), Y(1)\right) \perp T \vert X = x, \label{eqn:Unconf}
\end{equation}
\noindent where $\perp$ denotes independence of random variables.
\end{assumption}

The unconfoundedness assumption is usually not testable from the data; therefore, it must be accepted if it is reasonable to do so, given the underlying nature of the problem. In the case of credit line adjustments, the unconfoundedness assumption may hold if the adjustment is not influenced by unobserved factors related to both the treatments (i.e., credit line changes) and the outcome (expected profit). Furthermore, if numerous predictors are considered in the analysis, this assumption is strengthened. 

The definition of the conditional average treatment effect (CATE) which is the best estimator of the $ITE$ in terms of the mean squared error \cite{kunzel2019metalearners}, is as follows:

\begin{definition}
The conditional average treatment effect (CATE) of a subject with pretreatment features $X=x$, denoted by $\tau(x)$, is defined by the following equations:
\begin{equation}
    \tau(x):= \mathbb{E}\left(Y(1)-Y(0) \vert X=x\right).
\end{equation}
\end{definition}
In addition, if the data set satisfies Assumptions~\ref{assump:Overlap}-~\ref{assump:Unconf}, the $CATE$ can be estimated as 
\begin{align}
        \tau(x) &= \mathbb{E}\left(Y(1) \vert X=x\right)-\mathbb{E}\left(Y(0) \vert X=x\right) \notag\\
        &=\mathbb{E}\left(Y \vert T=1, X=x\right)-\mathbb{E}\left(Y \vert T=0, X=x\right).
\end{align}
\subsection{Causal Model for ITE Estimation}
\label{subsec:CausalModelsITE}
Below, we present several conventional causal models that can be employed for estimating $CATE$ and are relevant to our application of the proposed methodology, as we have benchmarked to them in the causal model selection stage.

\begin{itemize}
    \item \textbf{OLS/L1}: Here, a unique linear regression model is used to estimate the difference between the two potential outcomes; in this case, the treatment is considered an input to the model. This approach using any regression model is called \textit{direct modeling}, and it has been employed by \citet{johansson2016learning} and \citet{shalit2017estimating}.
    
    \item \textbf{Causal tree}: This is a regressor decision tree with split criteria for treatment effects instead of outputs. The most common splitting criterion is the heterogeneity in treatment effects; that is, the goal is to find splits that maximize the difference in outcomes between those treated and those in the control group per each subgroup. This methodology was created by \citet{athey2016recursive}.
    
    \item \textbf{Two-model approach:} In this approach, two separate regression models are used for each of the treated and control populations to estimate independently $\mu_1(x)=\mathbb{E}\left(Y(1) \vert X=x\right)=\mathbb{E}\left(Y \vert T=1, X=x\right)$ and $\mu_0(x)=\mathbb{E}\left(Y(0)\vert X=x\right)=\mathbb{E}\left(Y \vert T=0, X\right.$\\$\left.=x\right)$. This strategy is also known as the \textit{virtual twin} approach, and it has been used by \citet{lu2018estimating}. Particularly,  \textbf{OLS/L2} term denotes the two-model approach in which both regression models are linear.
    \item \textbf{X-Learner}: This method is an extension of the two-model approach, and it consists of the following three steps:
    \begin{enumerate}
        \item Using machine learning (ML) models, estimate the average outcomes $\mu_0(x)$ and $\mu_1(x)$.
        \item Input the level of treatment effects $D_k(1)$ and $D_j(0)$ for subjects $k$ in the treated group, with covariates $X_k(1)$ and subject $j$ in the control group with covariates $X_j(0)$, with  
        \begin{minipage}{0.4\textwidth}
\begin{align}
D_k(1)=Y_k(1)-\hat{\mu}_0\left(X_k(1)\right),
\end{align}
\end{minipage}
\hfill
\begin{minipage}{0.4\textwidth}
\begin{align}
D_j(0)=\hat{\mu}_1\left(X_j(0)\right)-Y_j(0),
\end{align}
\end{minipage}
\\
\\
\noindent and then estimate, applying ML models, $\tau_1(x)=\mathbb{E}\left(D(1) \vert X=x\right)$ and $\tau_0(x)=\mathbb{E}\left(D(0) \vert X=x\right)$.
        \item Finally, define the $CATE$ estimate as the weighted average: 
        \begin{equation}
            \hat{\tau}(x)=g(x)\hat{\tau}_0(x)+(1-g(x))\hat{\tau}_1(x),
        \end{equation}
        \noindent where $g$ is the propensity score, that is, $g(x)=\mathbb{P}\left(T=1 \vert X=x\right)$. 
    \end{enumerate}
    
    \item \textbf{R-Learner}: In this approach, we first estimate the treatment and control outcomes $\hat{\mu}(x)$ and the propensity score $g(x)$, using cross-validation. Afterwards, the estimate of $CATE$ is found by minimizing the following loss function:
    
    \begin{equation}
        \hat{L}(\tau(x))=\frac{1}{n}\sum_{i=1}^n \left[\left(Y_i-\hat{m}^{(-i)}(X_i)\right)-\left(T_i-g^{(-i)}(X_i)\right)\tau(X_i)\right]^2, 
    \end{equation}
    
    \noindent where $\hat{m}^{(-i)}(X_i)$ and $g^{(-i)}(X_i)$ are the predictions generated from the samples excluding the cross-validation fold to which the $i^{th}$ observation belongs.
    
    \item \textbf{Causal forest DML}: This approach combines the strengths of causal forests \citep{wager2018estimation, athey2019generalized}, with double machine learning \citep{chernozhukov2018double} to estimate heterogeneous treatment effects in causal inference tasks.
\end{itemize}

\subsection{Selection of Causal Models}
\label{subsec:ModelSelectionCM}
Selection of causal models is a challenge because it is not possible to compute the error in estimating the ground truth of the causal effects using real data, as we do not observe the counterfactuals. This condition is contrary to supervised learning, in which the standard strategy for model selection is to calculate the prediction error in a validation set and select the model with the lowest error.  

In the majority of studies, validation of causal models has been performed using synthetic data; in this case, parametric functions (e.g., polynomials or exponentials) have been used to simulate the outcomes of different treatments \citep{alaa2017bayesian, bica2020estimating}. However, it is crucial to consider that these conclusions have relied on strong assumptions regarding the response curves and may not have fully captured the intricate relationships present in real-world scenarios. 

Given this difficulty, \citet{alaa2019validating} have proposed a method for causal model selection using influence functions, a procedure used in robust statistics \citep{hampel2011robust} and efficiency theory \citep{robins2008higher} in which the loss of the causal model is estimated without the need to access counterfactual data. Here, we provide a brief explanation of this method, as we used it in our methodology for the selection of the best causal model. For more details, refer to \citet{alaa2019validating}.

In general, a causal model maps a set of observations $\{Z_i\}_{i=1}^{n}=\{(X_i, T_i, Y_i)\}_{i=1}^{n}$ (where $X_i, T_i$, and $Y_i$ represent the pretreatment features, the given treatment, and the outcome, respectively, for the $i^{th}$ observation in the observational data) to an estimate $\hat{\tau}$ of the $CATE$. We assume that the data comes from a probability distribution with parameter $\theta$, that is:
\begin{equation}
    Z_1, Z_2, \ldots, Z_n \sim \mathbb{P}_{\theta}, \label{eqn:Probabilitydistrib}
\end{equation}
\noindent with $\mathbb{P}_{\theta}$ in the family $\mathcal{P}=\{\mathbb{P}_{\theta'}: \theta' \in \Theta \}$, and with $\Theta$ the parameter space. The parameter $\theta$ is given by $\theta=\{\mu_0,\mu_1, g, \eta\}$ with $\eta=\mathbb{P}_{\theta}(X=x)$. The accuracy of the causal model is evaluated by the loss that occurred by its estimate; that is,
\begin{equation}
    l_{\theta}(\hat{\tau}):=\vert \vert \hat{\tau}(X)-\tau(X)\vert\vert_{\theta}^{2},
\end{equation}
\noindent with $\vert \vert f(X) \vert \vert_{\theta}^{2}:=\mathbb{E}_{\theta}[f^{2}(X)]$ for some function $f$. This measure has been termed the \textit{precision of estimating heterogeneous effects} (PEHE) \citep{hill2011bayesian}. The validation procedure fundamentally relies on understanding that if the model's loss is known under some known distribution $\mathbb{P}_{\tilde{\theta}}$ that is close to the true distribution $\mathbb{P}_{\theta}$, then $l(\theta)$ can be estimated via a Taylor expansion, as follows:
\begin{equation}
    l(\theta) \ \approx \ {\underbrace{l(\tilde{\theta})}_{\mathclap{\text{Plug-in estimate}}}} \ + \  {\underbrace{\dot{l}(\tilde{\theta})d(\theta-\tilde{\theta})}_{\mathclap{\text{Plug-in bias}}}}.\label{eqn:validationCausal}
\end{equation} 
According to Equation~\eqref{eqn:validationCausal}, the estimation of the required loss consists of two steps. In the first step, the observed data is used to find a synthetic distribution $\mathbb{P}_{\tilde{\theta}}$ (\textit{plug-in} distribution; that is, $\tilde{\theta}=\{\tilde{\mu}_0,\tilde{\mu}_1, \tilde{g}, \tilde{\eta}\}$) over which the loss $l(\tilde{\theta})$ is calculated. Second, the influential function $\dot{l}(\tilde{\theta})$ is found to correct from the bias using $\Tilde{\theta}$ instead of $\theta$ in the loss of the model (\textit{plug-in bias}). 

This procedure should be performed for all possible causal models. The goal is to select the best model, that is, the one with the minimum $PEHE$, considering the available data-set. As in supervised learning, no one causal model will perform best for all data-sets; in addition, with this state-of-the-art methodology, it has been shown that using only the factual information or only the plug-in step is not appropriate for selection of causal models. 

\section{Methodology}
\label{section:Methodology}
In this section, we propose a general methodology for providing recommendations, extending the manner in which the decisions are made by considering the estimation of treatment effects, accounting for their uncertainty, and including a prediction condition to refining the final policy.  

\subsection{Strategic Decision-Making for Contextual Impact With Dosage Treatment: Discretization Approach}
\label{subsec: StrategyCL}

Let us assume we aim to select the best action for every subject with pretreatment features $X$ among those not to be treated or receiving one treatment with a continuous-value range; that is, the possible treatment space $\mathcal{T}$ satisfies $\mathcal{T}=\{\beta \in \mathbb{R}:  \beta \in [\beta_{min}, \beta_{max}]\subseteq (0, \infty)\}$. This condition could represent the dosage of certain medications or the adequate increase factor for credit card limit modifications. 

As the decisions made will impact the context, rather than using only a
predictive model, we propose the involvement of causal learning inference. For this purpose, the observational data $\mathcal{S}^{o}$ is given by $\mathcal{S}^{o}=\{(X_i, T_i^{o}, Y_i): T_i^{o} \in \{0\} \cup \mathcal{T}\}_{i=1}^{n}$, with $X$ representing the pretreatment predictors; $T^{o}$ accounts for the given dosage received in the treatment or no treatment ($T^{o}=0$), and $Y$ indicates the observed continuous outcome (we have defined in our work that a greater value represents a better or more beneficial effect). For instance, in the case of recommendation of medicine doses, the desired outcome could be the probability of recovery after one year. For the credit card limit problem, its outcome could be defined as the generated expected profit.

The fulfillment of the overlap assumption is difficult to satisfy considering continuous-value treatments and observational data, a fact that has usually been ignored in previous studies. To make adequate comparisons between the outcomes under different treatments and the control, the treatment space is divided into $k$ treatment groups, according to the available real-world data; that is, we aim to select the best option among no treatment or the appropriate treatment dose $\beta$, with $\beta \in \mathcal{T}_D=\{\beta_1, \ldots, \beta_k\} \subseteq \mathcal{T}$. The coefficients $\beta_1, \ldots, \beta_k$ are determined based on a selected partition of the observable treatment dosages, denoted as $\mathcal{P}=\{t^{o}_1, \ldots, t^{o}_{k-1}\}$. Specifically, $\beta_1$ represents the average treatment dosage for individuals with a treatment dosage less than or equal to $t^{o}_1$. Similarly, $\beta_j$ (with $2\leq j \leq k-1$) represents the average treatment dosage for those with a treatment dosage between $t^{o}_{j-1}$ (exclusive) and $t^{o}_j$ (inclusive). Finally, $\beta_k$ denotes the average treatment dosage for individuals with a treatment dosage greater than $t^{o}_{k-1}$. Therefore, the observational data is transformed to $\tilde{\mathcal{S}}^{o}=\{(X_i, T_i, Y_i): T_i \in \{0\} \cup \mathcal{T}_D\}_{i=1}^{n}$, where the following conditions exist for all $i=1, \ldots, n$:
\begin{align}
T_i=0 & \Leftrightarrow T_i^{o}=0, \notag\\
T_i=\beta_1 & \Leftrightarrow T_i^{o}=\beta \textit{ and } \beta \leq t^{o}_1, \notag\\
T_i=\beta_j & \Leftrightarrow  T_i^{o}=\beta \textit{ and } t^{o}_{j-1}< \beta \leq t^{o}_{j}, \textit{ with } 2 \leq j \leq k-1, \notag \\
T_i=\beta_k & \Leftrightarrow T_i^{o}=\beta \textit{ and } \beta>t^{o}_{k-1}. \label{eqn:beta_factors}
\end{align}

To generate the final recommendation for each of the subjects, we decompose the general problem into the estimation of $k$ individual treatment effects $ITE_{\beta_j}$, defined as follows:
\begin{equation}
    ITE_{\beta_j}(X_i):=Y_i(\beta_j)-Y_i(0), \label{eqn:ITEk}
\end{equation}

\noindent for $j=0, \ldots, k$, with $\beta_0$ representing no treatment and $Y_i(\beta_j)$ and $ Y_i(0)$ representing the potential outcomes if the dose of the treatment was $\beta_j$, and if the subject had belonged to the control group for the $i-$th subject, respectively. According to Equation~\eqref{eqn:ITEk}, and without loss of generality, $ITE_{\beta_0}:=0$. Moreover, we have separated the decision-making problem into $k$ binary causal learning problems, aiming to estimate the $ITE$ given by Equation~\eqref{eqn:ITE_def} for each of them. In addition, as introduced in Subsection~\ref{subsec: Definitions and Assumptions}, the best estimator of the $ITEs$ is the estimator of the conditional average treatment effect, in this case, denoted by
\begin{equation}
    \tau_{\beta_j}(X_i):=\mathbb{E}\left[Y_i(\beta_j)-Y_i(0) \vert X=X_i\right].\label{eqn:CATEk}
\end{equation}
\subsection{Estimation of $ITE_{\beta_j}$ and Causal Learning Recommendation ($Re^{CL}$)}
\label{subsec:ReCL}
For the estimation of each $ITE_{\beta_j}$, for $j=1, \ldots, k$, different $k$ sets with overlap regions between the given treatment $\beta_j$ and the control are found given the real-world data. From this direction, the $k$ sets $\tilde{\mathcal{S}}^{o}_{\beta_j, 0}=\{(X_i,T_i, Y_i) \in \tilde{\mathcal{S}}^{o}: T_i \in \{0, \beta_j\} \}$, are used to train different propensity score models, denoted by:
\begin{equation}
 g_{\beta_j}(x)=\mathbb{P}(T=\beta_j \vert X=x), \label{eqn:PropensityModels}   
\end{equation}
\noindent for $j=1, \ldots, k$. Subsequently, these propensity models are used to generate the $k$ data sets $\tilde{\mathcal{S}}^{OV}_{\beta_j, 0}$ with overlapping regions, as follows:
\begin{equation}
 \tilde{\mathcal{S}}^{OV}_{\beta_j, 0}=\{(X_i,T_i, Y_i) \in \tilde{\mathcal{S}}^{o}_{\beta_j, 0}: 0<g_{\beta_j}(X_i)<1\}. \label{eqn:OverlapSets} 
\end{equation}
Different causal learning models are trained using $\tilde{\mathcal{S}}^{OV}_{\beta_j, 0}$ as the training set for $j=1, \ldots, k$, and the best causal model for estimating the individual treatment effect for each $\beta_j$ ($\mathcal{CM}_j$) is selected using the validation procedure described in Subsection~\ref{subsec:ModelSelectionCM}. 

Finally, to provide a recommendation over an unseen set with pretreatment predictors $\{X'_i\}_{i=1}^{m}$ (the number of recommendations, denoted as $m$, may differ from the number of observations, represented by $n$, which were used to train the causal models), the recommendation following causal learning (CL) criteria $Re^{CL}$ is as follows:
\begin{equation}
    Re^{CL}(X_i')= \arg \max_{l \in \mathcal{J}_i} \left\{\widehat{ITE}_{l}(X_i')\right\}, \label{eqn:ReCL}
\end{equation}
\noindent with $\mathcal{J}_i=\left\{l \in \{\beta_0,\ldots, \beta_k\}: \widehat{ITE}_{l}(X_i') \textit{ is defined}\right\}$, $\beta_0$ representing no treatment assignment (i.e., assignment to the control group) and the estimation of $ITE_{\beta_0}$ is provided by $\widehat{ITE}_{\beta_0}(X_i')=\widehat{\tau}_{\beta_0}(X_i'):=0$ and for $j= 1,\ldots, k$:
\begin{equation}
\widehat{ITE}_{\beta_j}(X_i')=\widehat{\tau}_{\beta_j}(X_i'):=\begin{cases} 
      \mathcal{CM}_j(X_i'), & 0<g_{\beta_j}(X_i')<1 \\
      \textit{Not defined}, & \textit{otherwise}.
   \end{cases}    \label{eqn:ITEjest}
\end{equation}
The definition given by Equation~\eqref{eqn:ITEjest} reveals that when propensity scores yield values of 0 or 1, the individual treatment effect is undefined. This fact occurs because in such cases, there is no prior history available for comparing treatment versus control. Additionally, in Equations~\eqref{eqn:ReCL} and~\eqref{eqn:ITEjest}, we are conducting a comparison of all potential individual treatment effects across multiple discrete treatment options. The objective is to identify the treatment option that provides the most beneficial causal impact when compared to the control group. The decision-making process employed in this analysis is characterized by caution. We exercise such caution by employing propensity score prediction models, which enable us to draw reliable conclusions from observations in the overlap regions. This approach ensures a robust assessment of treatment effects.

\subsection{Causal Learning and $CVaR$ Recommendation ($Re^{CL+CVaR}$): Improving the Decision-Making Process Using Uncertainty Measure of the ITE Estimations}
\label{subsec: StrategyCLCVaR}

When generating the recommendation $Re^{CL}$ according to Equation~\eqref{eqn:ReCL}, the estimation of the ITEs is considered for each subject; thus, a user can compare between certain treatments and the control. In such an approach, the uncertainty of the ITE predictions may lead a user to be indifferent to choosing between a few treatment options, as this uncertainty will, with a high chance, lead to confidence interval overlap. To avoid this fact, and to provide a more robust analysis, here we extend the methodology for selecting the best action outlined in Subsection~\ref{subsec:ReCL} by using an analogy of the conditional value-at-risk (CVaR) concept, which we have adapted for causal inference evaluation. The CVaR methodology  is extensively used in portfolio and actuarial management, but, it has not been evaluated thus far as a methodology to address the uncertainty of estimations regarding causal treatment impact.

For each subject $i \in \{1, \ldots, m\}$ for whom a recommendation should be provided and whose pretreatment predictors are  $X'_i$, first we generate the estimation of the $ITE_{\beta_j}$ for $j \in \{1, \ldots, k\}$, if applicable, following the procedure described in Subsection~\ref{subsec:ReCL}, in particular Equation~\eqref{eqn:ITEjest}. As an additional condition for suggesting no treatment or treatment with a certain dosage, we consider the uncertainty of the $ITE_{\beta_j}$ random variable for the observation $i$ ($ITE_{\beta_j}^i$), in case $\widehat{ITE}_{\beta_j}(X_i')$ is defined. For this purpose, we define the following two concepts:
\begin{definition}
\label{def:5.1}
The value-at-risk of a random variable (r.v) $X$ at the $100p\%$ level, denoted by $VaR_p\left(X\right)$, is the $100(1-p)$ percentile of the distribution of $X$.
\end{definition}
If the outcome of the treatment $Y$ is a continuous random variable, then $ITE_{\beta_j}^i$ is as well. Therefore, from Definition~\ref{def:5.1}, $VaR_p\left(ITE_{\beta_j}^i\right)$ is the value that satisfies the following condition:
\begin{equation}
    \mathbb{P}\left(ITE_{\beta_j}^i \leq VaR_p\left(ITE_{\beta_j}^i\right)\right)=1-p. \label{eqn:PercentileVaR}
\end{equation}

To understand this concept, we see positive values of the $ITE_{\beta_j}^i$ random variable as gains, while negative values represent losses. Consequently, the concept of value-at-risk tells us that with $100p\%$ confidence, the probability that the losses are less than $VaR_p\left(ITE_{\beta_j}^i\right)$ is ($1-p$). The \textit{value-at-risk} identifies an extreme value of a loss given a certain confidence level; however, to obtain substantial information regarding the average loss value, that is, regarding the tail of the distribution of $ITE_{\beta_j}^i$, the notion of \textit{conditional value-at-risk} of $ITE_{\beta_j}^i$ is introduced in the following definition:  
\begin{definition}
\label{def:5.2}
The conditional value-at-risk of $X$ at the $100p\%$ level, denoted by $CVaR_p\left(X\right)$ is determined by
\begin{align}
    CVaR_p\left(X\right)&= \mathbb{E}\left[X \vert X \leq VaR_p\left(X\right) \right]=\frac{\int_{-\infty}^{VaR_p\left(X\right)}xf_{X}(x)dx}{F_{X}\left(VaR_p\left(X\right)\right)}, \label{eqn:CVaR}
\end{align}
where $f_{X}(\cdot)$ and $F_{X}(\cdot)$ are the density and distribution functions, respectively, of $X$.
\end{definition}

Definitions~\ref{def:5.1} and~\ref{def:5.2} are similar to those provided in \cite[pp. 43-45]{klugman2012loss}; however, in contrast, losses are indicated by negative values in our work. Because $CVaR_p\left(ITE_{\beta_j}^i\right)$ not only captures the size of potential losses but also indicates the expected magnitude of these losses beyond the $VaR_p\left(ITE_{\beta_j}^i\right)$, we employ this measure of uncertainty to gain a more complete recommendation than $Re^{CL}$, which considers not only the causal learning estimations but also their uncertainty given by the $CVaR$ forecast. This new recommendation $Re^{CL+CVaR}$ over an unseen set with pretreatment predictors $\{X'_i\}_{i=1}^{m}$ is assigned by the following equation:
\begin{equation}
    Re^{CL+CVaR}(X_i')= \arg \max_{l \in \mathcal{J}_i} \left\{CVaR_{p}\left(ITE_{l}^i\right)\right\}, \label{eqn:ReCLCvar}
\end{equation}
where $\mathcal{J}_i=\left\{l \in \ \{\beta_0,\ldots, \beta_k\}: \widehat{ITE}_{l}(X_i') \textit{ is defined}\right\}$, $CVaR_{p}\left(ITE_{\beta_0}^i\right):=0$ and for calculating the \textit{conditional value-at-risk} $CVaR_{p}\left(ITE_{\beta_j}^i\right)$, the bootstrap distribution of the prediction made using the selected causal models $\widehat{ITE}_{\beta_j}(X_i')$ given in Equation~\eqref{eqn:ITEjest} is generated.  Subsequently, an approximation of the $CVaR_p$ is obtained as the average of the losses that exceed the $VaR_p$.

\subsection{Causal Learning, $CVaR$, and Forward-Looking Recommendation ($Re^{CL+CVaR+FL}$): Refining the Recommendation Using Prediction Criteria}
\label{subsec: StrategyCLCVaRFL}

Although the recommendation $Re^{CL+CVaR}$ presented in Equation~\eqref{eqn:ReCLCvar} is more complete than $Re^{CL}$ from Equation~\eqref{eqn:ReCL}, as the former considers not only the estimation of the individual treatment effects but also the potential extreme negative values of the increase decision, we can further enrich the final recommendation by including a forward-looking prediction criterion that allows us to confirm or reject the decision provided by Equation~\eqref{eqn:ReCLCvar}. This approach is suitable for cases in which the defined outcome could be observable before and after treatment. For instance, the probability of survival before and after treatment has been administered in certain medical settings or the expected profit before and after a credit limit change or maintenance for credit line modification can be easily measured.

Because we are aiming to select between no treatment or a specific-dosage treatment that entails the best causal treatment impact, we impose that the recommendation of increase assigned by $Re^{CL+CVaR}$ is ratified if the pretreatment outcome is less than the predicted post-treatment outcome; this condition occurs when, after selecting the treatment, the expected future output surpasses the past observable output. To determine the future outcome that will depend on the treatment, we should use the best-performing machine learning model. Therefore, this new recommendation $Re^{CL+CVaR+FL}$ over an unseen set with pretreatment predictors $\{X'_i\}_{i=1}^{m}$ is assigned by

\begin{equation}
    Re^{CL+CVaR+FL}(X_i'):=\begin{cases} 
      \beta_0, &  Re^{CL+CVaR}(X_i')= \beta_0 \textit{ or } Y_R \geq \hat{Y}_P  \\
       Re^{CL+CVaR}(X_i'), & \textit{otherwise},
   \end{cases}, \label{eqn:ReCLCvarFL}
\end{equation}

\noindent with $Y_R$ being the observable pretreatment outcome and $\hat{Y}_P$ being the predicted future outcome, given the corresponding treatment $Re^{CL+CVaR}(X_i')$.

\begin{figure}[ht]
\centering
	\includegraphics[width=0.9\textwidth]{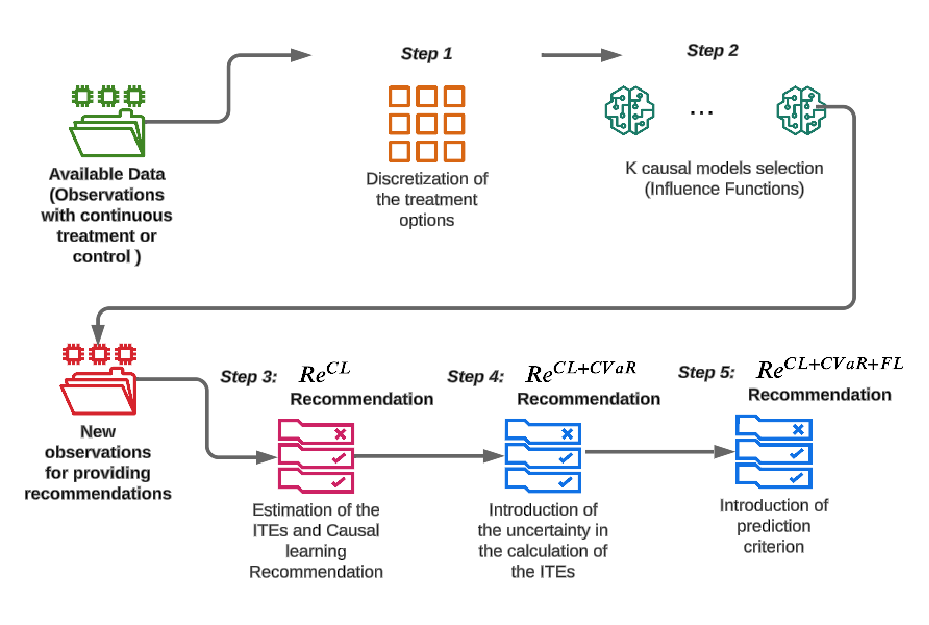}
	\caption{Process for treatment recommendation.}
	\label{fig:process_met}
\end{figure}

The final process has been summarized in Figure~\ref{fig:process_met}.
\section{Results}
In this section, we describe how the methodology presented in Section~\ref{section:Methodology} was applied to the specific problem of credit limit adjustments. For this purpose, we used information about credit card holders of a fintech company. Specifically, we employed financial data from customers who had mantained an active credit card for at least six months. In this period, some of these customers had received an increase in their credit line. Our goal was to use this historical behavioral data to develop a data-driven method to recommend credit line modifications. 

\subsection{Credit Limit Adjustment as a Causal Learning Problem}

\begin{figure}[ht]
\centering
	\includegraphics[width=0.8\textwidth]{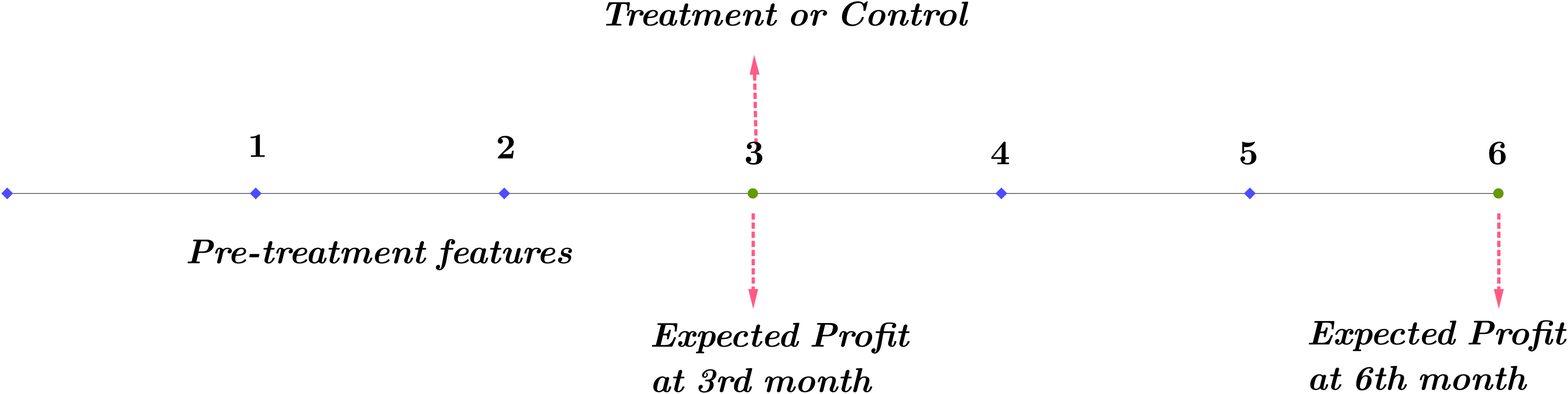}
	\caption{Periodicity of the decision for credit line adjustment and measurement of the impact.}
	\label{fig:periodiocity-Impact}
\end{figure}

Based on the observational data at hand, our approach to measuring the treatment impact involved considering the pretreatment data, which included the outcome variable, before determining whether a cardholder should be given the treatment (an increase in a certain percentage of their original credit limit) or the control (no increase). Additionally, we considered the outcome variable following the action of increasing the credit line or keeping it constant. In our specific case, the pretreatment outcome variable corresponded to the expected profit in the month the treatment was implemented (third month), and the post-treatment outcome variable represented the expected profit three months after the modification of the credit line (sixth month). Figure~\ref{fig:periodiocity-Impact} represents the times of measurement of the pretreatment variables and the outcome after the treatment or control actions.

For calculating the expected profits before and after treatment, we calculated the expected revenue minus the provisions, in which the latter term refers to the capital (money) required to cover potential losses stemming from nonpayments (defaults):
\begin{align}
    \mathbb{E}(Profit^i_n) &= \mathbb{E}(Revenue^i_n)-Provisions^i_n\notag\\
    &= Int^i \cdot Bal^i_n \cdot (1-PD^i_n)-PD^i_n \cdot LGD \cdot EAD^i_n, \label{eqn:EProfit}
\end{align}
\noindent where $i$ represents the information regarding the $i^{th}$ credit-card holder and $n$ stands for the time in which the expected profit $\mathbb{E}(Profit^i_n)$ is being calculated (that is, $n=3, 6$ for the third and sixth months, respectively). Additionally, the following definitions relate to the $i^{th}$ customer:
\begin{itemize}
    \item  $Int^i$ is the monthly interest rate.
    \item $Bal^i_n$ is the outstanding balance at time $n$.
    \item $PD^i_n$ is the probability of defaulting at time $n$. This probability was computed using a regulatory model specific to their locality. This model considered various factors, such as the payment percentage, credit card usage, and the frequency of missed payments. 
    \item $LGD$ is the loss given default, which is assumed to be constant for all customers and time-independent according to international banking regulations \citep{basel2010basel}.
    \item $EAD^i_n$ is the exposure to default at time $n$.
\end{itemize}

In Equation~\eqref{eqn:EProfit}, for determining the provisions, we refer to the guidelines of the Basel III Accord, in which the expected loss or provision  -- the necessary reserve of liquid assets that should be maintained to cover projected losses due to defaults -- is estimated by the multiplication of $PD^i_n, LGD$, and $EAD^i_n$. For computing the exposure at default $EAD^i_n$, to capture the unique characteristic of revolving products, it is essential to incorporate a model that accounts for the tendency of borrowers nearing default to utilize a higher proportion of their remaining credit limit. To address this issue, the regulations require using a value termed the credit conversion factor (CCF). The CCF considers the proportion of the remaining credit limit that is expected to be utilized if a borrower defaults. Specifically,
\begin{equation}
    EAD^i_n = Bal^i_n+CCF \cdot \left(L^i_n-Bal^i_n\right),
\end{equation}
\noindent where $L^i_n$ is the credit limit for the $i^{th}$ customer at month $n$. The $CCF$ is determined by analyzing the data regarding individuals who defaulted, following the approach given by \citet{tong2016exposure}.

\begin{figure}[ht]
\centering
    \textbf{Histogram of increase factors}\par\medskip
	\includegraphics[width=0.5\textwidth]{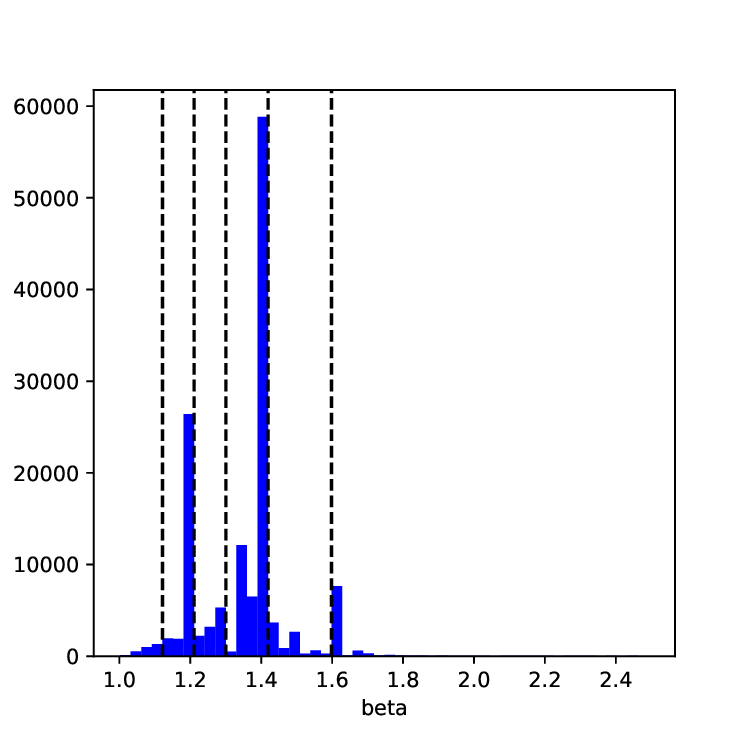}
	\caption{Partition of the factors of increase.}
	\label{fig:Retrospective-Prospective}
\end{figure}

 Our dataset consisted of information regarding 257,640 customers active in the six months of observation, of whom 140,180 had received an increase in the third month. We considered only credit line increases of less than or equal to 2.5 times compared to the previous credit line, as 99.3\% of the credit card portfolio satisfied this condition. First, we divided the factor of increase into six groups $\{\beta_1, \beta_2, \ldots, \beta_6\}$, following the mechanism described in Equation~\eqref{eqn:beta_factors} and according to the characteristics of the past decisions. The partition of the treatment dosage is shown in Figure~\ref{fig:Retrospective-Prospective}, and the description of the variables used in our work is presented in Table~\ref{tab:Variables-Explanation}.
\begingroup
\renewcommand{\arraystretch}{1.2} 
\begin{table}[ht]

\centering
\caption{Variable definitions.}
\resizebox{\textwidth}{!}{%
\begin{tabular}{@{}p{\textwidth}@{}}
\multicolumn{1}{c}{\textbf{Explanation of Variables}} \\ \midrule
1. Bureau score calculated when the credit card was assigned. \\ \hline
2. Estimated income of the customer when the credit card was obtained. \\ \hline
3. Interest rate that was imposed on each customer. \\ \hline
4. Number of months with the credit card. \\ \hline
5. Credit limits in the third and sixth months. \\ \hline
6. The monthly average outstanding balance at cut-off dates during the entire history of the \ \ credit card. \\ \hline
7. Monthly average consumption during the credit card time records. \\ \hline
8. Balances at the cut-off dates during the first, second, and third months. \\ \hline
9. Total consumption in each of the three first months using the fintech application. \\ \hline
10. Total consumption in each of the three first months outside the fintech application. \\ \hline
11. Average number of credit card fees per each of the three first months. \\ \hline
12. Maximum number of credit card fees per each of the three first months. \\ \hline
13. Minimum number of credit card fees per each of the three first months. \\ \hline
14. Total number of transactions in each of the first three months. \\ \hline
15. Maximum number of unpaid bills across the credit card account's lifetime. \\ \hline
16. Number of unpaid statements in the period of the first three months. \\ \hline
17. Expected profit in the third and sixth months. \\ \hline
\end{tabular}}
\label{tab:Variables-Explanation}
\end{table}
\endgroup

\begin{figure}[H]
\centering
\includegraphics[width=\textwidth]{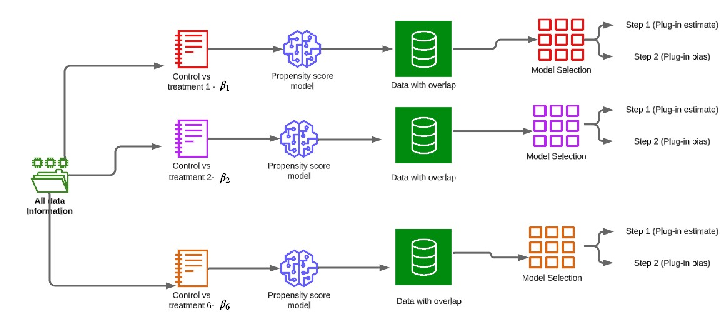}
	\caption{Selection of causal models for credit adjustment decisions.}
	\label{fig:methodology1}
\end{figure}

Subsequently, we constructed six distinct datasets following the dynamics of each account evolution. This approach enabled us to conduct a comprehensive comparison between customers in the control group, in which the credit limit remained unchanged ($T = 0$), and those in the treatment group. Specifically, the treatment group included individuals subject to credit limit increases ($T = \beta_i$) according to one of six predetermined factors ($i = 1, \ldots, 6$). The primary aim of this design was to ensure the crucial overlap assumption. Namely, we constructed six propensity score models $g_{\beta_j}$ to generate six data sets with an overlap region between treatment and control $ \tilde{\mathcal{S}}^{OV}_{\beta_j, 0}$, as proposed in Equations~\eqref{eqn:PropensityModels} and ~\eqref{eqn:OverlapSets}, respectively. After obtaining these datasets with explicit overlap, we could select causal models based on the methodology described in Subsection~\ref{subsec:ModelSelectionCM}. Figure~\ref{fig:methodology1} presents the strategy used in selecting the causal model for deciding between keeping the credit limit constant (the control) and increasing it by a given factor (the treatment). 

The results from performing the causal model selection per each of the treatment factors $\beta_1, \ldots, \beta_6$, using influence functions, have been presented in Appendix~\ref{App:CasualModelSelection}.
For our particular dataset, we deemed it necessary to calculate approximations of the second-order influence functions of the PEHE functional, which resemble second-order derivatives of a function in standard calculus (the loss function was approximated via a generalized Taylor expansion). This decision was made because, in contrast with the case study in \citep{alaa2019validating}, using only the first-order influence functions did not provide consistent results. To validate our findings, we further computed the approximations of the third-order influence functions and obtained consistent results. Based on these findings, the optimal causal models for each treatment increase factor were as follows: For $\beta_1$ and $\beta_3$, a causal forest DML model yielded the best results; for $\beta_2$ and $\beta_5$, a causal tree model was most effective; for $\beta_4$, a two-model approach using random forest models as regressors proved successful; and finally, for $\beta_6$, the most suitable model was an $X$-learner with random forests as regressors.

Above, we have presented the way in which the causal model selection was made by following the methodology described in  Subsection~\ref{subsec:ModelSelectionCM}. However, in addition to performing model selection, an essential step before deploying any model in practice is to perform an external evaluation. In contrast with supervised learning, in which we would have selected a test set with which to measure the held-out error of the predictions, in employing counterfactual learning, we could not observe all potential outcomes given a test set. Therefore, for external evaluation, we would have needed to deploy the model in practice and measure whether the potential outcomes improved according to the recommendation provided by the causal model (A/B testing) \citep{bica2021real}. Because this option was not feasible in our situation, we performed a back-testing evaluation by comparing the output results corresponding to different scenarios, specified as follows:
\begin{enumerate}
    \item[I.] Cases in which the increase factor suggested by the given strategy (GS, derived from some specific criteria (e.g., $Re^{CL}$)), aligned with the company's past policy (CPP); 
    \item[II.] Situations in which both the CPP and GS recommended an increase, but with different factors for modifying the credit line;
    \item[III.] Cases in which the GS called for an increase in the credit line while the CPP advised keeping the credit limit unchanged;
    \item[IV.] Instances in which the GS recommended keeping the credit limit while the CPP suggested raising it;
    \item[V.] Instances in which the GS and CPP matched in recommending to keep the credit line constant.
\end{enumerate}
To execute this evaluation, we selected a test set that had not been used during the training of the causal models. This test set included data regarding 131,559 credit card holders and their financial information during the same six months. The decisions provided by the $CPP$ are described in Table~\ref{tab:CPP}.

\begin{table}[H]
\centering
\caption{Distribution of the past decisions made by the company}
\begin{tabular}{cr}
\textbf{$CPP$}                       & \textbf{Number of Customers} \\ \midrule
\textbf{Keeping credit line constant} & 89,834                        \\
Increase $\beta_4$                   & 23,189                        \\
Increase $\beta_2$                   & 9,070                         \\
Increase $\beta_3$                   & 3,174                         \\
Increase $\beta_6$                   & 2,943                         \\
Increase $\beta_5$                   & 2,515                         \\
Increase $\beta_1$                   & 834                          \\ 
\end{tabular}
\label{tab:CPP}
\end{table}

Under $CPP$, in the test data, approximately 32\% of the credit card holders received an increase. On the other hand, the factual outputs under the $CPP$ are shown in Table~\ref{tab:FactualResults}, in which $EP_R$ and $EP_P$ represent the expected profit at the third and sixth months, respectively (i.e., before and after treatment, when applicable), $EP_R/L_R$ is the ratio between the expected profit and the credit limit at the third month, and $EP_P/L_R$, the expected profit at the sixth month over the credit limit at the third month. All monetary results in this work have been expressed in generic monetary units (\textit{Mu}). This masking is done to protect confidential information from the corporate partner. The numbers indicated are real values scaled by a constant, random, positive, factor for all values.
\begin{table}[ht]
\centering
\caption{Factual results under $CPP$.    ``Avg" denotes average, and ``per cust." represents per customer. Expected profits measured in scaled monetary units (\textit{Mu}).}
\begin{tabular}{lccccc}
CPP (Factual) & \multicolumn{1}{p{1.7cm}}{\centering Sample\\ Size.} &  \multicolumn{1}{p{1.7cm}}{\centering Avg\\ $EP_P$ \\ per cust} &  \multicolumn{1}{p{1.7cm}}{\centering Avg\\ $EP_P/L_R$\\ per cust.\\ (\%)} &  \multicolumn{1}{p{1.7cm}}{\centering Avg $EP_R$;\\ Avg $EP_P$\\ per cust.} &  \multicolumn{1}{p{1.7cm}}{\centering Avg\\ $EP_R/L_R$;\\ Avg \\ $EP_P/L_R$\\ per cust. (\%)} \\ \midrule
Keep constant      & 89,834                                                      & 1.221                                                             & 0.90                                                                        & 3.346; 1.221                                                                    & 1.73; 0.90                                                                                       \\
Increase      & 41,725                                                      & 3.339                                                             & 2.08                                                                        & 3.194; 3.339                                                                    & 1.85; 2.08                                                                                       \\
All decisions & 131,559                                                     & 1.893                                                             & 1.28                                                                        & 3.298; 1.893                                                                    & 1.77; 2.08                                                                                       \\ 
\end{tabular}
\label{tab:FactualResults}
\end{table}

In the following subsections, we present the evaluation results using different strategies, including those defined in Subsections~\ref{subsec: StrategyCL} to \ref{subsec: StrategyCLCVaRFL}. We began with the best-performing policy, which incorporated all concepts: estimation of individualized treatment effects, calculation of the $CVaR$, and inclusion of the forward-looking criterion ($Re^{CL+CVaR+FL}$). Subsequently, we explored policy performance by relaxing the criteria used to verify that $Re^{CL+CVaR+FL}$ surpassed policies generated using less-stringent conditions. In all results that follow, the $CVaR$ was calculated using a level of $0.95$, that is, using $CVaR_{0.95}$.

\subsection{Results Under $Re^{CL+CVaR+FL}$ Recommendation}
In this specific scenario, after preliminary experiments, we found that the model that produced the best policy evaluation results in generating the forward-looking criterion described in Subsection~\ref{subsec: StrategyCLCVaRFL} was a multilayer perceptron (MLP). In addition, its out-of-sample error in terms of the root mean squared error was 5.95 \textit{Mu} (monetary units), which was 46\% of the target standard deviation (i.e., the expected profit in the sixth month).

According to Table~\ref{tab:CLCVaRFL}, this policy recommended a credit line increase to approximately 54\% of the customers in the test dataset, exceeding the 32\% corresponding with the recommendation provided by $CPP$; thus, 28,943 more customers would have received an increase in their credit line if this policy had been followed. In Table~\ref{tab:EvaluationCLCVaRFL}, we display the evaluation of this suggestion following the different scenarios (I.-V).
\begin{table}[H]
\centering
\caption{Distribution of recommendations originating from $Re^{CL+CVaR+FL}$ suggestion.}
\begin{tabular}{cr}
\textbf{$Re^{CL+CVaR+FL}$}                       & \textbf{Number of Customers} \\ \midrule
\textbf{Maintenance of credit line} & 60,480                      \\
Increase $\beta_4$                   & 31,914                       \\
Increase $\beta_2$                   & 22,549                         \\
Increase $\beta_6$                   & 15,890                         \\
Increase $\beta_5$                   & 633                          \\ 
Increase $\beta_3$                   & 90                         \\ 
Increase $\beta_1$                   & 3                          \\ 
\end{tabular}
\label{tab:CLCVaRFL}
\end{table}

The information in Table~\ref{tab:EvaluationCLCVaRFL} provides valuable insights into the average expected return ($EP_P/L_R$) per customer when the policy suggested by $Re^{CL+CVaR+FL}$ aligned with the company's chosen policy for credit line increases. The observed average expected return of $2.96\%$ was significantly higher than the $2.60\%$ return when the factor of increase differed. Moreover, using this strategy, the customers who would have received an increase but did not receive one based on the past policy produced a better-expected return of $1.86\%$, in comparison to the $0.73 \%$ return related to those who would have had their credit lines maintained under $Re^{CL+CVaR+FL}$, but to whom the company decided to offer a limit increase. Conversely, when both policies coincided with keeping the credit lines constant, the expected return of $0.03\%$ was the smallest among the five scenarios. This observation demonstrates the effectiveness of $Re^{CL+CVaR+FL}$ in identifying less-profitable customers and recommending credit limit increases. Additionally, it can be deduced from Table~\ref{tab:EvaluationCLCVaRFL} that the expected profit return from the customers for whom the recommendation $Re^{CL+CVaR+FL}$ was to increase the credit limit was $2.51\%$. This value surpassed the expected profit return of $2.08\%$ for customers who had been granted an increase based on the company's past decisions, as indicated in Table~\ref{tab:FactualResults}.
\begin{table}[ht]
\centering
\caption{Evaluation results derived from $Re^{CL+CVaR+FL}$ recommendation. ``Avg" denotes average and ``per cust." represents per customer. Expected profits have been measured in scaled monetary units (\textit{Mu}).}
\resizebox{\textwidth}{!}{%
\begin{tabularx}{1.06\textwidth}{llccccc}
\multicolumn{1}{p{1.7cm}}{\centering \ \\ \textbf{Recommendation}}& \multicolumn{1}{p{1.7cm}}{\centering \ \\ \textbf{CPP (Factual)}} & \multicolumn{1}{p{1.7cm}}{\centering \ \\ \textbf{Sample\\ size} } &  \multicolumn{1}{p{1.7cm}}{\centering \ \\ \textbf{Avg\\ $EP_P$ \\ per cust.}} &  \multicolumn{1}{p{1.7cm}}{\centering \centering \ \\ \textbf{Avg\\ $EP_P/L_R$\\ per cust.\\ (\%)}} &  \multicolumn{1}{p{1.7cm}}{\centering \centering \ \\ \textbf{Avg $EP_R$;\\ Avg $EP_P$\\ per cust.}} &  \multicolumn{1}{p{1.7cm}}{\centering \textbf{Avg\\ $EP_R/L_R$;\\ Avg \\ $EP_P/L_R$\\ per cust. (\%)}} \\ \midrule
Increase same factor & Increase & 10,480 & 4.839 & 2.96 & 2.416; 4.839 & 1.64; 2.96  \\
Increase different factor & Increase & 17,701 & 5.103 & 2.60 & 2.983; 5.103 & 1.62; 2.60  \\
Increase & Keep constant  & 42,898 & 3.976 & 1.86 & 3.404; 3.976  & 1.70; 1.86\\
Keep constant & Increase  & 13,544 & -0.128 & 0.73 & 4.072; -0.128 & 2.32; 0.73\\
Keep constant & Keep constant & 46,936 &-1.298 & 0.03 & 3.293;-1.298 & 1.75; 0.03 \\ 
\end{tabularx}
}
\label{tab:EvaluationCLCVaRFL}
\end{table}

To avoid potential biases arising from differences in expected profits and credit limit amounts, our analysis focused on the expected return rather than the raw expected profit. Additionally, we have provided information regarding average expected profits and expected returns before and after treatment to provide a comprehensive view of the results.
\subsection{Results Built on the $Re^{CL+CVaR}$ Proposal}
When we relaxed the previous criteria for making the final recommendation of credit line adjustment and used only the causal learning framework together with the estimation of the uncertainty of predictions accounted by the $CVaR_{0.95}$ calculation (i.e., $Re^{CL+CVaR}$), we found a significant difference from $CPP$ and $Re^{CL+CVaR+FL}$. In this case, based on Table~\ref{tab:CLCVaR}, the proportion of customers who would have received an increase was $78\%$. This finding indicates that $Re^{CL+CVaR}$ carried a higher level of risk compared to $Re^{CL+CVaR+FL}$.
\begin{table}[H]
\centering
\caption{Distribution of recommendations based on the $Re^{CL+CVaR}$ recommendation}
\begin{tabular}{cr}
\textbf{$Re^{CL+CVaR}$}                       & \textbf{Number of Customers} \\ \midrule
\textbf{Maintenance of credit line} & 29,058                      \\
Increase $\beta_4$                   & 43,093                       \\
Increase $\beta_2$                   & 36,133                         \\
Increase $\beta_6$                   & 22,178                         \\
Increase $\beta_5$                   & 976                        \\
Increase $\beta_3$                   & 117                         \\
Increase $\beta_1$                   & 4                          \\ 
\end{tabular}
\label{tab:CLCVaR}
\end{table}

In terms of the policy evaluation in the different scenarios, we can conclude from Table~\ref{tab:EvaluationCLCVaR} that this strategy still produced competent results; for instance, the average expected return from the customers for whom the $Re^{CL+CVaR}$ factor of increase matched the company's past decision was $2.83\%$, which was greater than $2.38\%$ return that corresponded with those raises that did not agree with the new suggestion. However, the expected return for increases on which the past policy and the new suggestion concurred was less than the $2.96\%$ return obtained using the more-complete recommendation given by $Re^{CL+CVaR+FL}$. Thus, we can conclude that the inclusion of the predictive condition enhanced the suggestion of credit line modification that had been generated using only  counterfactual analysis and uncertainty measurement through $CVaR$ computation. 
\begin{table}[ht]
\centering
\caption{Evaluation results derived from $Re^{CL+CVaR}$ recommendation. ``Avg" denotes average, and ``per cust." represents per customer. Expected profits have been expressed in scaled monetary units (\textit{Mu}).}
\resizebox{\textwidth}{!}{%
\begin{tabularx}{1.1\textwidth}{llccccc}
\multicolumn{1}{p{1.7cm}}{\centering \ \\ \textbf{Recommendation}}& \multicolumn{1}{p{1.7cm}}{\centering \ \\ \textbf{CPP (Factual)}} & \multicolumn{1}{p{1.7cm}}{\centering \ \\ \textbf{Sample\\ size} } &  \multicolumn{1}{p{1.7cm}}{\centering \ \\ \textbf{Avg\\ $EP_P$ \\ per cust.}} &  \multicolumn{1}{p{1.7cm}}{\centering \centering \ \\ \textbf{Avg\\ $EP_P/L_R$\\ per cust.\\ (\%)}} &  \multicolumn{1}{p{1.7cm}}{\centering \centering \ \\ \textbf{Avg $EP_R$;\\ Avg $EP_P$\\ per cust.}} &  \multicolumn{1}{p{1.7cm}}{\centering \textbf{Avg\\ $EP_R/L_R$;\\ Avg \\ $EP_P/L_R$\\ per cust. (\%)}} \\ \midrule
Increase same factor & Increase& 14,159  & 4.454  & 2.83 & 2.929; 4.454 & 2.02; 2.83 \\
Increase different factor & Increase& 24,965 & 4.585 & 2.38 & 3.758; 4.585  & 1.90; 2.38   \\
Increase & Keep constant & 63,377 & 3.438 & 1.73 & 4.066; 3.438 & 1.94; 1.73 \\
Keep constant & Increase & 2,601 & -14.700  & -4.8  & -0.780; -14.700  & 0.41; -4.8     \\
Keep constant & Keep constant & 26,457 & -4.090 & -1.08 & 1.621; -4.090  & 1.22; -1.08  \\ 
\end{tabularx}
}
\label{tab:EvaluationCLCVaR}
\end{table}
\subsection{Results Stemming From $Re^{CL}$ Prescription}
If we had relied solely on the recommendation based on the estimation of individualized treatment effects ($Re^{CL}$), the credit limit would have been increased for 93\% of the customers in the test dataset. However, this approach exhibits a key limitation, as it fails to account for uncertainty in these estimations. As a result, it provides flexible and risk-tolerant suggestions that may pose potential harm to the company's profit and, more importantly, could lead to overindebtedness of customers in the future. Furthermore, as depicted in Table~\ref{tab:CL}, 32\% of the customers with suggestions for increase would have received the higher factor of credit line raise.

Table~\ref{tab:EvaluationCL} shows that $Re^{CL}$ provided more appropriate recommendations to keep credit lines constant even when the company had chosen to increase them, as these customers produced a negative average expected profit return of $-12.37\%$. However, the average expected profit return of $2.75\%$ for this policy when the factors of increase were aligned with CPP did not outperform the previous results obtained from $Re^{CL+CVaR}$ and $Re^{CL+CVaR+FL}$, which were $2.83\%$ and $2.96\%$ returns, respectively.
\begin{table}[H]
\centering
\caption{Allocation of recommendations based on $Re^{CL}$ recommendation.}
\begin{tabular}{cr}
\textbf{$Re^{CL}$}                       & \textbf{Number of Customers} \\ \midrule
\textbf{Maintenance of credit line} & 8,727                      \\
Increase $\beta_6$                   & 39,097                       \\
Increase $\beta_4$                   & 30,711                         \\
Increase $\beta_2$                   & 28,609                          \\ 
Increase $\beta_3$                   & 11,432                          \\
Increase $\beta_1$                   & 10,607                          \\
Increase $\beta_5$                   & 2,376                          \\
\end{tabular}
\label{tab:CL}
\end{table}
\begin{table}[ht]
\centering
\caption{Evaluation of results derived from $Re^{CL}$ recommendation. ``Avg" denotes average, and ``per cust." represents per customer. Expected profits have been expressed in scaled monetary unit (\textit{Mu}).}
\resizebox{\textwidth}{!}{%
\begin{tabularx}{1.12\textwidth}{llccccc}
\multicolumn{1}{p{1.7cm}}{\centering \ \\ \textbf{Recommendation}}& \multicolumn{1}{p{1.7cm}}{\centering \ \\ \textbf{CPP (Factual)}} & \multicolumn{1}{p{1.7cm}}{\centering \ \\ \textbf{Sample\\ size} } &  \multicolumn{1}{p{1.7cm}}{\centering \ \\ \textbf{Avg\\ $EP_P$ \\ per cust.}} &  \multicolumn{1}{p{1.7cm}}{\centering \centering \ \\ \textbf{Avg\\ $EP_P/L_R$\\ per cust.\\ (\%)}} &  \multicolumn{1}{p{1.7cm}}{\centering \centering \ \\ \textbf{Avg $EP_R$;\\ Avg $EP_P$\\ per cust.}} &  \multicolumn{1}{p{1.7cm}}{\centering \textbf{Avg\\ $EP_R/L_R$;\\ Avg \\ $EP_P/L_R$\\ per cust. (\%)}} \\ \midrule
Increase same factor & Increase&   11,380 & 4.331 & 2.75& 2.872; 4.331 & 1.93; 2.75\\
Increase different factor &  Increase & 29,194 & 4.403 & 2.39 & 3.628; 4.403 & 1.82; 2.39 \\
Increase & Keep constant & 82,258 & 2.999 & 1.64 & 3.958; 2.999 & 1.98; 1.64  \\
Keep constant & Increase & 1,151 & -33.480 & -12.37 & -4.642; -33.480 & -1.32; -12.37 \\
Keep constant & Keep constant & 7,576 & -18.087 & -7.09 & -3.297; -18.087 & -1.01; -7.09 \\ 
\end{tabularx}}
\label{tab:EvaluationCL}
\end{table}

Incorporating the computation of $CVaR$ into the recommendation alongside the initial estimation of ITEs ($Re^{CL}$) not only allowed us to adopt a more risk-averse approach but also produced better results in suggesting credit line increases. Based on the data from Tables~\ref{tab:CLCVaR} and \ref{tab:CL}, it can be deduced that the expected profit return from the customers who received a credit line raise was $1.92\%$ under $Re^{CL}$ and $2.04\%$ under $Re^{CL+CVaR}$.
 
 \subsection{Results Based Only on Prediction Conditions}
In this subsection, we evaluate the performance of a policy generated solely using a supervised learning approach. The recommendation was derived using a regression model to predict the future expected profit, with the factor of increase as one of the input features. The final decision involved selecting the factor that produced the maximum future expected profit, if it exceeded the previous expected profit, and recommending keeping the limit constant otherwise. The following results were obtained using the same regression model used for generating $Re^{CL+CVaR+FL}$, specifically the multilayer perceptron (MLP) model.

According to Table~\ref{tab:PredictionDistribution}, when employing the supervised approach, approximately $73\%$ of the customers represented in the dataset would have received a credit line increase, with the majority being granted the highest factor of credit line raise. However, this result still indicates a less risk-averse approach compared to the policy generated by $Re^{CL+CVaR+FL}$. Additionally, based on Table~\ref{tab:EvaluationPrediction}, the average expected profit return when the factor of increase coincided with CPP was $2.62\%,$ whereas it was $2.69\%$ when these factors did not agree. Moreover, the average expected profit return over the increases with this policy was $2.12\%,$ which was lower than that provided by $Re^{CL+CVaR+FL}$.
\begin{table}[H]
\centering
\caption{Distribution of suggestions based only on the prediction condition.}
\begin{tabular}{cr}
\textbf{Prediction Recommendation}                       & \textbf{Number of Customers} \\ \midrule
\textbf{Constant credit line} & 34,938                      \\
Increase $\beta_6$                   & 59,496                       \\
Increase $\beta_5$                   & 14,176                         \\
Increase $\beta_1$                   & 13,027                          \\ 
Increase $\beta_3$                   & 5,028                          \\
Increase $\beta_2$                   & 3,033                         \\
Increase $\beta_4$                   & 1,861                          \\
\end{tabular}
\label{tab:PredictionDistribution}
\end{table}
\begin{table}[H]
\centering
\caption{Evaluation of results derived from following only the prediction criteria. ``Avg" denotes average, and ``per cust." represents per customer. Expected profits have been expressed in scaled monetary units (\textit{Mu}).}
\resizebox{\textwidth}{!}{%
\begin{tabularx}{1.08\textwidth}{llccccc}
\multicolumn{1}{p{1.7cm}}{\centering \ \\ \textbf{Recommendation}}& \multicolumn{1}{p{1.7cm}}{\centering \ \\ \textbf{CPP (Factual)}} & \multicolumn{1}{p{1.7cm}}{\centering \ \\ \textbf{Sample\\ size} } &  \multicolumn{1}{p{1.7cm}}{\centering \ \\ \textbf{Avg\\ $EP_P$ \\ per cust.}} &  \multicolumn{1}{p{1.7cm}}{\centering \centering \ \\ \textbf{Avg\\ $EP_P/L_R$\\ per cust.\\ (\%)}} &  \multicolumn{1}{p{1.7cm}}{\centering \centering \ \\ \textbf{Avg $EP_R$;\\ Avg $EP_P$\\ per cust.}} &  \multicolumn{1}{p{1.7cm}}{\centering \textbf{Avg\\ $EP_R/L_R$;\\ Avg \\ $EP_P/L_R$\\ per cust. (\%)}} \\ \midrule
Increase same factor & Increase&  2,097 & 4.396 & 2.62& 1.738; 4.396
& 1.31; 2.62\\
Increase different factor &  Increase & 30,607& 4.685& 2.69 & 2.813; 4.685 & 1.76; 2.69\\
Increase & Keep constant & 63,917 & 3.420 & 1.85 & 3.038; 3.420 & 1.79; 1.85 \\
Keep constant & Increase & 9,021 & -1.475 & 0.00 & 4.824; -1.475
 & 2.28; 0.00   \\
Keep constant & Keep constant & 2,5917 & 0.899 & 0.76 & 3.487; 0.899 & 1.77; 0.76  \\ 
\end{tabularx}
}
\label{tab:EvaluationPrediction}
\end{table}
\section{Discussion}
Our analysis has revealed significant insights into the effectiveness of various strategies for adjusting credit lines. Initially, we observed that relying solely on a supervised learning approach for credit line adjustment was inadequate. Despite its simplicity and ease of understanding, this method failed to yield optimal results. For instance, when the increase factor aligned with the CPP policy, the average expected profit return was 2.62\%, marginally lower than the 2.69\% return observed when these factors diverged.

Furthermore, policies derived from estimating treatment effects ($Re^{CL}$) led to a high proportion of credit line increases (93\%), making this strategy overly aggressive for practical application. This approach also overlooked the uncertainty associated with estimations of ITEs. By integrating the conditional value-at-risk ($CVaR$) into our policy recommendations ($Re^{CL+CVaR}$), we introduced a more conservative stance, resulting in a 78\% rate of increase. However, even this adjusted approach fell short of delivering optimal outcomes.

Our methodology, which incorporated ITE estimations, their uncertainty (via $CVaR$), and a predictive criterion (asserting that future expected returns should surpass past ones), emerged as the most prudent, with a moderate proportion of increases of 54\%. Crucially, it also secured the highest expected return, marking it as the superior strategy. These findings strongly advocate for the empirical testing of our comprehensive approach in practical scenarios.

\section{Conclusion}

In this research, we have proposed a novel strategy for performing recommendations involving multiple treatment options. Our approach prioritizes the satisfaction of the overlap assumption, a critical requirement for drawing reliable conclusions regarding the outcomes of different treatments and controls. To address this aspect, we have developed a methodology that involves generating distinct data-sets with overlap before proceeding with causal model training.


Our proposal is focused on causal inference model selection using influence functions. Notably, to the best of our knowledge, this technique was applied to banking data for the first time through this study, necessitating the calculation of second- and third- order influence function approximations. Subsequently, we answered three proposed research questions, assessing the effectiveness of the uncertainty measure and the inclusion of an adequate predictive condition. While the problem we addressed, limit-setting in financial services, has commonly been within the territory of traditional predictive machine learning, the fact that our objective was to estimate potential outcomes that were not directly observable made this problem much better suited for causal models.

We applied our proposed methodology to the task of credit limit modification, advocating for the use of alternative data-driven techniques to address this issue. This problem holds significant importance not only in traditional banking but also in emerging business environments such as fintech companies. In particular, we aimed to estimate the individual treatment effect arising from a specific treatment (increasing the credit limit) compared to a control action (keeping the credit limit constant). The treatment's outcome was defined as the expected future profit, reflecting two competing objectives, namely maximizing individual revenue and minimizing customer provision, and including the concept of credit conversion factor to design a solution that was accurate in the context of credit card portfolio risk management.

Our findings demonstrate that relying purely on either pure machine learning or counterfactual predictions (\textbf{RQ3}) is insufficient for recommending optimal treatments. Instead, by considering the uncertainty measure of the estimations through the conditional value-at-risk (\textbf{RQ1}) and imposing an appropriate prediction condition (\textbf{RQ2}), we were able to formulate a policy that was less risky and more profitable than all individual approaches.

In conclusion, our research has contributed a valuable framework for multitreatment recommendations, emphasizing the importance of addressing the overlap assumption and employing advanced methodologies for causal inference with uncertainty in the estimations. By successfully applying our methodology to credit limit adjustment, we have highlighted its potential to improve decision-making. Ultimately, our study has underscored the significance of combining multiple criteria to derive more reliable and efficient policies for complex decision-making scenarios.

\section*{Acknowledgment}
The authors gratefully acknowledge the support of the Natural Sciences and Engineering Research Council of Canada (NSERC) [Discovery Grants RGPIN-2020-07114 and RGPIN-2019-06586]. This work was funded, in part, with funding from the Canada Research Chair Program [CRC-2018-00082], and was enabled, in part, by support provided by Compute Ontario (\url{https://www.computeontario.ca}) and the Digital Research Alliance of Canada (\url{https://alliancecan.ca}).

\appendix
\section{Causal Models Selection}
\label{sec:appendix}
During the training stage of the causal models, we employed \texttt{econml} library  \cite{battocchi2019econml} for the  Causal Forest DML model, and \texttt{causalml} library \cite{chen2020causalml} for the Causal Tree, Two-model, $X$-learner, and $R$-learner approaches.
\begingroup
\renewcommand{\arraystretch}{1.5} 
\begin{table}[H]
\centering
\caption{Result of the final $\sqrt{PEHE}$ over 5-validation folds and its standard deviation ($SD$), using second order influence functions. $RF$ and $XGB$ denote the use of random forest models and xgboost models as regressors in each of the different approaches $T$ (Two model approach), $X$ (X-learner) or $R$ (R-learner). Measures are given in scaled monetary units (\textit{Mu}).}
\resizebox{\textwidth}{!}{%

\begin{tabularx}{1.15\textwidth}{lcccccc} 
\multicolumn{1}{p{1cm}}{\centering \textbf{Causal \\  Model}}      & \multicolumn{1}{p{1.5cm}}{\centering \textbf{Treatment \\ $\beta_1$}} & \multicolumn{1}{p{1.5cm}}{ \centering \textbf{Treatment \\ $\beta_2$}} & \multicolumn{1}{p{1.5cm}}{\centering \textbf{Treatment \\ $\beta_3$}}  & \multicolumn{1}{p{1.5cm}}{\centering \textbf{Treatment \\ $\beta_4$}}  & \multicolumn{1}{p{1.5cm}}{\centering \textbf{Treatment \\ $\beta_5$}} & \multicolumn{1}{p{1.5cm}}{\centering \textbf{Treatment \\ $\beta_6$}}\\ \hline
                           & \multicolumn{6}{c}{\centering \textbf{$\sqrt{PEHE}\pm 
                           SD$}}                                                                                                                                                                                                                      \\ \hline
\textbf{OLS/L2}            &  20.542$\pm$9.247                                                       & 7.175$\pm$0.732  & 7.546$\pm$0.835      & 5.641$\pm$0.415 & 8.283$\pm$1.105                                                          & 13.760$\pm$6.901                                                                                                           \\ \hline
\textbf{OLS/L1}            & 13.365$\pm$5.592 & 6.955$\pm$0.768   & 8.002$\pm$0.403 & 5.754$\pm$0.640                                                          & 7.964$\pm$0.690                                                          & 10.735$\pm$2.019                                                                                       \\ \hline
\multicolumn{1}{p{1cm}}{\centering \textbf{Causal \\ Forest DML}} & \multicolumn{1}{p{2cm}}{\centering \textbf{ \\ 11.886$\pm$5.520}}                                               & \multicolumn{1}{p{2cm}}{\centering \ \\ 5.803$\pm$0.653}                                                         & \multicolumn{1}{p{2cm}}{\centering \textbf{ \\ 6.544$\pm$0.640}}  & \multicolumn{1}{p{2cm}}{\centering \ \\ 5.154$\pm$0.709}                                                        & \multicolumn{1}{p{2cm}}{\centering \ \\6.890$\pm$1.127}                                                      & \multicolumn{1}{p{2cm}}{\centering \ \\ 11.141$\pm$3.983}                                                 \\ \hline
\textbf{Causal Tree}       & 14.981$\pm$3.754 & \textbf{5.546$\pm$0.783}  & 7.422$\pm$0.508   & 4.793$\pm$0.799  & \textbf{6.810$\pm$0.575}   & 9.839$\pm$2.535\\ \hline
\textbf{T-RF}              & 18.993$\pm$4.557 & 6.648$\pm$0.811 & 8.353$\pm$0.866
 &\textbf{4.281$\pm$0.412} & 7.308$\pm$1.064 & 11.019$\pm$1.178
                                                      \\ \hline
\textbf{T-XGB}             & 12.282$\pm$5.070                                                        & 6.403$\pm$0.677                                                          & 7.503$\pm$0.318 & 5.327$\pm$0.628                                                          & 7.410$\pm$0.678                                                          & 10.062$\pm$2.013                                                        \\ \hline
\textbf{X-RF}              & 12.159$\pm$5.027                                                       & 5.582$\pm$0.464                                                          & 6.840$\pm$0.627     & 5.213$\pm$0.759                                                          & 7.050 $\pm$1.286                                                        & \textbf{9.218  $\pm$1.859}                                                          \\ \hline
\textbf{X-XGB}             & 12.071$\pm$5.111                                                         & 6.362$\pm$0.695                                                         & 7.412$\pm$0.391 & 5.483$\pm$0.676                                                          & 7.380$\pm$0.645                                                          & 10.064$\pm$2.195                                                      \\ \hline
\textbf{R-RF}              & 30.720$\pm$6.716                                                         & 11.612$\pm$0.514                                                         & 19.872$\pm$9.893  & 18.429$\pm$6.530 & 64.125$\pm$25.941 & 59.389$\pm$9.664                                                              \\ \hline
\textbf{R-XGB}             & 15.654$\pm$5.156& 6.403$\pm$0.678  & 9.215$\pm$0.632  & 5.827$\pm$0.571  & 9.878$\pm$0.662 & 13.40$\pm$1.242\\ 
\end{tabularx}
}
\label{tab:CausalModelSelection_2}
\end{table}
\endgroup

\begingroup
\renewcommand{\arraystretch}{1.5} 
\begin{table}[H]
\centering
\caption{Result of the final $\sqrt{PEHE}$ over 5-validation folds and its standard deviation ($SD$), using third order influence functions. Measures are given in scaled monetary units (\textit{Mu}).}
\resizebox{\textwidth}{!}{%
\begin{tabularx}{1.15\textwidth}{ccccccc} 
\multicolumn{1}{p{1cm}}{\centering \textbf{Causal \\  Model}}      & \multicolumn{1}{p{1.5cm}}{\centering \textbf{Treatment \\ $\beta_1$}} & \multicolumn{1}{p{1.5cm}}{ \centering \textbf{Treatment \\ $\beta_2$}} & \multicolumn{1}{p{1.5cm}}{\centering \textbf{Treatment \\ $\beta_3$}}  & \multicolumn{1}{p{1.5cm}}{\centering \textbf{Treatment \\ $\beta_4$}}  & \multicolumn{1}{p{1.5cm}}{\centering \textbf{Treatment \\ $\beta_5$}} & \multicolumn{1}{p{1.5cm}}{\centering \textbf{Treatment \\ $\beta_6$}}\\ \hline
                           & \multicolumn{6}{c}{\centering \textbf{$\sqrt{PEHE}\pm 
                           SD$}}                                                                                                                                                                                                                      \\ \hline
\textbf{OLS/L2}            &  20.558$\pm$9.248
                                                      & 7.176$\pm$0.735  & 7.554
$\pm$0.844      & 5.641$\pm$0.415   & 8.301$\pm$1.112
                                                         & 13.857$\pm$6.748                                                                                                           \\ \hline
\textbf{OLS/L1}            & 13.390$\pm$5.624 & 6.957$\pm$0.770  & 8.013$\pm$0.415 & 5.754$\pm$0.639                                                          & 7.988$\pm$0.662                                                          & 10.828$\pm$1.799                                                                                       \\ \hline
\multicolumn{1}{p{1cm}}{\centering \textbf{Causal \\ Forest DML}} & \multicolumn{1}{p{2cm}}{\centering \textbf{ \\ 11.911$\pm$5.551}}                                               & \multicolumn{1}{p{2cm}}{\centering \ \\ 5.804$\pm$0.657 }                                                         & \multicolumn{1}{p{2cm}}{\centering \textbf{ \\ 6.554$\pm$0.642}}  & \multicolumn{1}{p{2cm}}{\centering \ \\ 5.155$\pm$0.709}                                                        & \multicolumn{1}{p{2cm}}{\centering \ \\6.917$\pm$1.094}                                                      & \multicolumn{1}{p{2cm}}{\centering \ \\ 11.255$\pm$3.721}                                                 \\ \hline
\textbf{Causal Tree}       & 14.996$\pm$3.800 & \textbf{5.547$\pm$0.780}  & 7.422
$\pm$0.508    & 4.793$\pm$0.798  & \textbf{6.834$\pm$0.559}   & 9.943$\pm$2.269 \\ \hline
\textbf{T-RF}              & 19.012$\pm$4.576  & 6.649$\pm$0.810 & 8.362$\pm$0.862
  &\textbf{4.281$\pm$0.412} & 7.327$\pm$1.074 & 11.090$\pm$1.147                                                       \\ \hline
\textbf{T-XGB}             & 12.309$\pm$5.104                                                         & 6.405$\pm$0.679                                                         &7.514$\pm$0.330 & 5.327$\pm$0.627                                                          & 7.435 $\pm$0.654                                                         & 10.160$\pm$1.793                                                           \\ \hline
\textbf{X-RF}              & 12.187$\pm$5.059                                                         & 5.583$\pm$0.465                                                          & 6.840$\pm$0.627     & 5.213$\pm$0.759                                                         & 7.078$\pm$1.240                                                         & \textbf{9.327$\pm$1.592}                                                          \\ \hline
\textbf{X-XGB}             & 12.100$\pm$5.145                                                         & 6.364$\pm$0.697                                                          & 7.412$\pm$0.391 & 5.483$\pm$0.675                                                          & 7.406$\pm$0.610                                                         & 10.168$\pm$1.954                                                      \\ \hline
\textbf{R-RF}              & 29.264$\pm$3.892                                                        & 11.373$\pm$0.836                                                         & 20.896$\pm$6.884  & 17.536$\pm$4.677 & 62.803$\pm$28.496 & 58.950$\pm$6.311                                                              \\ \hline
\textbf{R-XGB}             & 15.677$\pm$5.180 & 6.405$\pm$0.679   & 9.247$\pm$0.622  & 5.827$\pm$0.553  & 9.901$\pm$0.657 &13.466$\pm$1.080\\ 
\end{tabularx}
}
\label{tab:CausalModelSelection_3}
\end{table}
\endgroup

\end{document}